\documentclass[conference]{IEEEtran}
\IEEEoverridecommandlockouts

\usepackage{cite}
\usepackage{amsmath,amssymb,amsfonts}
\usepackage{algorithmic}
\usepackage{graphicx}
\usepackage{textcomp}
\usepackage{xcolor}
\usepackage{algorithm}
\usepackage{algorithmic}
\usepackage{graphicx}
\usepackage{float}
\usepackage{subcaption}
\usepackage{tikz}
\usepackage{pgfplots}
\usepackage{pgfplotstable}
\usepackage{tikz}
\usetikzlibrary{decorations.markings}
\pgfplotsset{compat=1.18}
\usepackage{booktabs}
\usepackage[xindy]{glossaries}  

\newacronym[description={Markov Decision Process}]{mdp}{MDP}{Markov decision process}
\newacronym{rl}{RL}{Reinforcement Learning}
\newacronym{il}{IL}{Imitation Learning}
\newacronym{hrl}{HRL}{Hierarchical Reinforcement Learning}
\newacronym{ika}{IKA}{Institute for Automotive Engineering}
\newacronym{hp}{HP}{Highway Pilot}
\newacronym{sad-pilot}{SAD-Pilot}{Scenario-based Automated Driving Pilot}
\newacronym{sad-rl}{SAD-RL}{Scenario-based Automated Driving Reinforcement Learning}
\newacronym{lca}{LCA}{Lane Change Assistant}
\newacronym{mpc}{MPC}{Model Predictive Controller}
\newacronym{ttc}{TTC}{Time To Collision}
\newacronym{wes}{WES}{With Easy Scenarios}
\newacronym{ads}{ADS}{Automated Driving Systems}
\newacronym{ddt}{DDT}{Dynamic Driving Task}
\newacronym{oedr}{OEDR}{Object and Event Detection and Response}
\newacronym{rwth}{RWTH University}{Rheinisch-Westfälische Technische Hochschule University}

\makeglossaries

\begin{document}

\title{Scenario-Based Hierarchical Reinforcement Learning for Automated Driving Decision Making}

\author{
\IEEEauthorblockN{M. Youssef Abdelhamid}
\IEEEauthorblockA{Machine Learning and Reasoning Chair\\
RWTH Aachen University \\
youssef.abdelhamid@rwth-aachen.de}
\and
\IEEEauthorblockN{Lennart Vater}
\IEEEauthorblockA{Institute for Automotive Engineering\\
RWTH Aachen University \\
lennart.vater@ika.rwth-aachen.de}
\and
\IEEEauthorblockN{Zlatan Ajanovic}
\IEEEauthorblockA{Machine Learning and Reasoning Chair\\
RWTH Aachen University \\
zlatan.ajanovic@ml.rwth-aachen.de}
}


\maketitle

\begin{abstract}
Developing decision-making algorithms for highly automated driving systems remains challenging, since these systems have to operate safely in an open and complex environments.
Reinforcement Learning (RL) approaches can learn comprehensive decision policies directly from experience and already show promising results in simple driving tasks.
However, current approaches fail to achieve generalizability for more complex driving tasks and lack learning efficiency.
Therefore, we present \gls{sad-rl}, the first framework that integrates Reinforcement Learning (RL) of hierarchical policy in a scenario-based environment.
A high‑level policy selects maneuver templates that are evaluated and executed by a low‑level control logic.
The scenario-based environment allows to control the training experience for the agent and to explicitly introduce challenging, but rate situations into the training process.
Our experiments show that an agent trained using the \gls{sad-rl} framework can achieve safe behaviour in easy as well as challenging situations efficiently.
Our ablation studies confirmed that both HRL and scenario diversity are essential for achieving these results.
\end{abstract}


\section{Introduction}

Automated Driving Systems (ADS) aim to enhance passenger comfort and road traffic safety \cite{10.3389/fhumd.2021.669030}. Developing systems capable of operating autonomously in complex traffic environments poses significant challenges \cite{kremer2020scenario}. These systems must not only reliably perceive their surroundings but also plan actions, predict the behavior of other road users, and react to unexpected behaviors, such as a lead vehicle braking harshly. Ensuring safe and effective decision-making in such scenarios is crucial for avoiding collisions and maintaining safety.

\gls{rl} has emerged as a promising approach for handling complex decision-making in ADS \cite{silver2016mastering}. \gls{rl} enables an agent to learn optimal behaviors through trial-and-error interactions with the environment. However, \gls{rl} methods often suffer from slow learning rates, sample inefficiency, and unsafe exploratory actions during training. Hierarchical Reinforcement Learning (HRL) can potentially address these issues by structuring decision-making into high-level options and low-level policies, thus improving learning efficiency and safety \cite{9210154}.

Despite recent progress, existing approaches to automated driving systems typically address only a subset of the three essential requirements. First, safety (R1) is critical, as the system must reliably prevent traffic accidents and hazardous situations. Second, sample efficiency (R2) is important to ensure that the system can learn effective driving policies within a reasonable amount of training time. Third, generalizability (R3) requires the system to transfer learned behaviors across various driving conditions, adapting to new scenarios without extensive retraining.

This paper proposes a framework that combines RL learning of hierarchical policy with a structured, scenario-based training environment to meet all three requirements. Unlike many existing methods that rely on random traffic simulations, which may lack consistency and challenge, our approach leverages systematically designed scenarios to expose the agent to a diverse and progressively difficult set of situations. These scenarios simulate critical events that encourage efficient and targeted learning.

A review of existing methods shows that while some approaches meet certain requirements, no single method fully satisfies all three. For example, DQN-attention \cite{liu2022combining} and ASAP-RL \cite{wang2023efficient} address safety and sample efficiency but fail to satisfy the generalizability. In contrast, BC-SAC approach \cite{lu2023imitation} demonstrates safety and generalizability but it remains sample-inefficient as it requires training on large datasets ranging from 40k to 2 million scenarios in order to achieve robustness. This research aims to address these gaps by developing a comprehensive approach that meets all specified requirements, thus contributing to the advancement of safe and efficient automated driving systems.

The main contributions of this paper are:
\begin{enumerate}
    \item Integration of HRL with structured scenario-based simulation for automated driving, resulting in the proposed \gls{sad-rl} framework that enables maneuver abstraction and safe low-level control (Sec. \ref{ch:methodology}) 
    \item Development of scenario datasets that combine synthetically generated critical scenarios and real-road extracted ones, facilitating the exposure to safety-relevant events and promoting generalizability (Sec. \ref{sc:scenarios})
    \item Empirical demonstration of \gls{sad-rl} improved sample efficiency, safety and generalizability, including ablation studies that confirm the necessity of both HRL and scenario-based learning for robust transferable policy learning (Sec. \ref{sc:experiments} and \ref{sc:discussion}).
\end{enumerate}

\section{Related Work}
\label{ch:RW}

Decision-making in ADS has been addressed through a range of approaches, including rule-based systems, planning algorithms, \gls{il} and \gls{rl}. Although existing approaches have strengths, they often fail to satisfy all the key requirements of safety, sample efficiency and generalizability.

\subsection{Rule-based and Planning-based Approaches}
\label{sc:rulebased}
Rule-based methods define hand-engineered rules inspired by traffic laws like adhering to speed limits, lane centering and collision avoidance. These rules can ensure safety but they lack flexibility in novel or uncertain situations.
Planning-based methods generate trajectories from the current state to the goal state using search algorithms. For example, Fraichard et al. \cite{fraichard1993dynamic} used A* search algorithm for planning in dynamic environments. Ajanovic et al. \cite{ajanovic2018search} introduced geometric spatio-temporal representation of different traffic elements, rules, and agents and employed A* search to general driving scenarios. More recently, Saraoglu et al. \cite{saraoglu2023minimax} considers interaction between vehicles and used Minimax planning to find the safe trajectories even under worst-case assumptions for other traffic participants. However, they struggle with generalizability to unpredictable interactions and real-time computation requirements.

\subsection{Imitation Learning}
\label{sc:immitationlearning}
Imitation learning (IL) aims to mimic expert behavior, with Behavioral Cloning (BC) being a common method that minimizes the difference between predicted and expert actions. Early work like ALVINN~\cite{pomerleau1988alvinn} applied BC for end-to-end driving from sensor inputs. However, BC struggles with generalization due to limited data diversity. More recent approaches such as Symphony~\cite{igl2022symphony} use Generative Adversarial Imitation Learning (GAIL) to address covariate shift by aligning learned policies with expert trajectories through adversarial training, incorporating hierarchical structure and evaluating policy diversity and realism.

\subsection{Reinforcement Learning}
\label{ssc:RL}

\gls{rl} is widely used in robotics and automated systems. Researchers have explored combining \gls{rl} with other methods, such as imitation learning. For instance, Lu et al. \cite{lu2023imitation} developed BC-SAC (Behavioral Cloning - Soft Actor-Critic), which combines IL and \gls{rl} to enhance safety and reliability in challenging scenarios. BC-SAC uses a dual actor-critic architecture similar to Twin Delayed Deep Deterministic Policy Gradients (TD3) and Soft Actor-Critic (SAC). It includes an actor-network $\pi(a|s)$, a double Q-critic network Q(s, a), and a target double Q-critic network \( \overline{Q}(s, a) \), with a Transformer observation encoder for vehicle states, road-graph points, traffic signals, and route goals. BC-SAC is trained on 100k miles of real-world trajectories stratified by difficulty. Performance is measured via failure rates across test slices of increasing difficulty, showing significant gains over IL- and RL-only baselines.

Other research has addressed safety issues in \gls{rl}. For example, Alshiekh et al. \cite{alshiekh2018safe} introduced shielding to prevent unsafe actions and penalize mistakes, thus enhancing the safety of \gls{rl} approaches. SafetyNet \cite{vitelli2022safetynet} employed a Machine-Learned planner trained on expert driving data and introduced a fallback trajectory to handle potential collisions, achieving a 95\% reduction in collisions compared to ML planners alone. Combining rule-based approaches with automated methods like IL or \gls{rl} has proven effective in improving safety and performance.

\section{Methodology}
\label{ch:methodology}

\begin{figure}[H]
    \centering
    \includegraphics[width=0.48\textwidth]{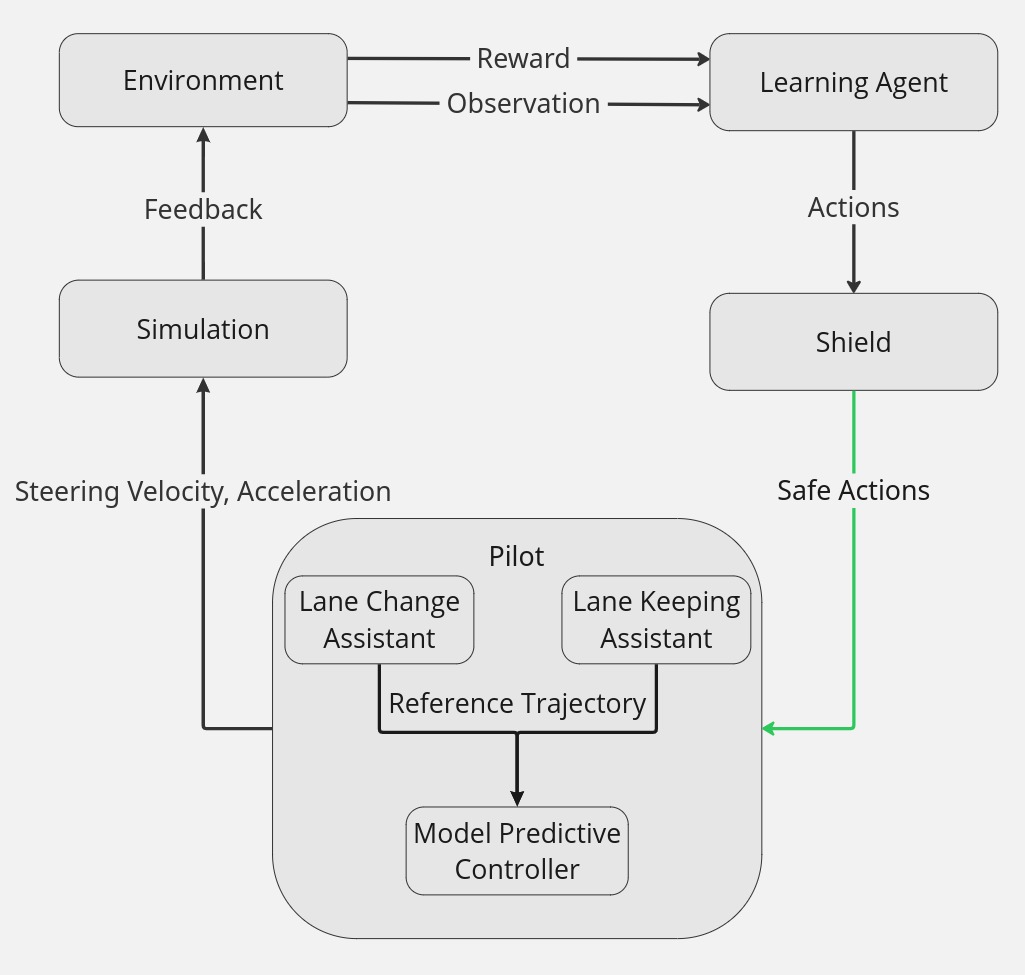}
    \caption{\gls{sad-rl} system architecture showing the hierarchical interaction between the Learning Agent, Shield, and Pilot, including its low-level execution modules and interaction with the simulation environment.}
    \label{fig:sad_rl_pipeline}
\end{figure}

This section describes the components of the framework used in this work. \gls{sad-rl}, as shown in Fig.~\ref{fig:sad_rl_pipeline}, integrates a hierarchical reinforcement learning agent, a scenario-based simulation environment, and a shielding mechanism to ensure safe and efficient training. The system uses both synthetic and real-world extracted scenarios. It applies HRL to decouple high-level decisions from low-level executions. 

\subsection{Simulation}
\label{sc:simulation}
We use a CommonRoad-based simulation toolchain\cite{Althoff2017a}, chosen for its lightweight structure and modular scenario description. Unlike platforms such as CARLA or SUMO, CommonRoad avoids unnecessary complexities for this research (e.g., dense sensor models) while allowing full control over vehicle and road layout. Scenarios are defined in XML or JSON and rendered with annotated vehicles and goals. The simulation is open-loop, where non-ego vehicles follow predefined trajectories. In the simulation, the ego vehicle interacts with a simplified observation space including road layout, vehicle state $(s_x, s_y, v, \delta, \psi)$ and surrounding traffic. Termination occurs under five conditions: goal reached, collision, timeout, standstill, or offroad.

\subsection{Scenarios}
\label{sc:scenarios}
The scenarios in this research are generally highway scenarios that are divided into synthetically generated and real-road extracted types. This section discusses the method used to generate synthetic scenarios and the method of extraction of real-road scenarios from highway recorded data.

\subsubsection{Synthetically Generated Scenarios}
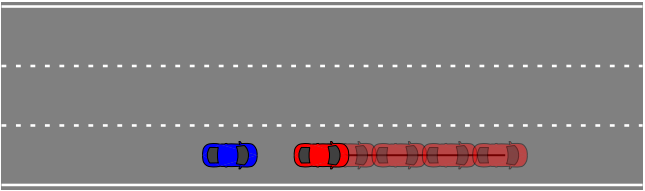
\begin{figure}[h]
    \centering
    \resizebox{\linewidth}{!}{%
        \usetikzlibrary{decorations.markings}

\tikzset{
    car_top/.pic={
    \filldraw [black, fill=#1, very thin] plot[smooth, black, tension=.7, very thin] 
    coordinates {(2.3248,0) (2.2967,0.4058) (2.0941,0.7925)(1.7847,0.9958) (1.1874,1.0261) (0.9497,1.0108)}
    .. controls (0.9124,1.143) and (0.8648,1.2352) .. (0.7463,1.1888) .. controls (0.7628,1.0703) and (0.7832,1.0089) .. (0.8202,0.952) .. controls (0.6029,0.9902) and (0.1582,1.0079) .. (-0.2536,0.9816) .. controls (-0.3325,1.0605) and (-0.4289,1.0459) .. (-0.5136,0.9787) .. controls (-0.6947,0.9904) and (-0.8232,0.9758) .. (-0.94,0.9524)
    plot[smooth, tension=.7,very thin] coordinates {(-0.94,0.9524)(-1.1951,1.0129) (-1.8081,0.9951) (-2.1686,0.8328) (-2.3894,0.4889) (-2.4286,0) };

    \fill [fill=#1,very thin](-2.4286,0)--(-0.94,0.9524)--(2.3248,0)--(-0.94,-0.9524);

    \filldraw  [black, fill=#1, very thin] plot[smooth, black, tension=.7,very thin] 
    coordinates {(2.3248,0) (2.2967,-0.4058) (2.0941,-0.7925)(1.7847,-0.9958) (1.1874,-1.0261) (0.9497,-1.0108)}
    .. controls (0.9124,-1.143) and (0.8648,-1.2352) .. (0.7463,-1.1888) .. controls (0.7628,-1.0703) and (0.7832,-1.0089) .. (0.8202,-0.952) .. controls (0.6029,-0.9902) and (0.1582,-1.0079) .. (-0.2536,-0.9816) .. controls (-0.3325,-1.0605) and (-0.4289,-1.0459) .. (-0.5136,-0.9787) .. controls (-0.6947,-0.9904) and (-0.8232,-0.9758) .. (-0.94,-0.9524)
    plot[smooth, tension=.7,very thin] coordinates {(-0.94,-0.9524)(-1.1951,-1.0129) (-1.8081,-0.9951) (-2.1686,-0.8328) (-2.3894,-0.4889) (-2.4286,0) };

    \draw [smooth cycle, black, fill=darkgray,very thin](0.52,0.75) .. controls (0.7,0.78) and (1,0.82) .. (1.2,0.85) .. controls (1.7,0.5) and (1.7,-0.5) .. (1.2,-0.85) .. controls (1,-0.82) and (0.7,-0.78) .. (0.52,-0.75) .. controls (0.7,-0.25) and (0.7,0.25) .. (0.52,0.75);
    \draw [smooth cycle, black, fill=darkgray,very thin](-1.06,0.67) .. controls (-1.33,0.68) and (-1.57,0.67) .. (-1.8,0.68) .. controls (-2.1,0.27) and (-2.1,-0.27) .. (-1.8,-0.68) .. controls (-1.57,-0.67) and (-1.33,-0.68) .. (-1.06,-0.67) .. controls (-1.16,-0.3) and (-1.16,0.3) .. (-1.06,0.67);
    }
}

\begin{tikzpicture}
\fill [gray] (-7,-2.3) rectangle (7,1.8);
\draw [white, ultra thick] (-7,1.7)--(7,1.7);
\draw [loosely dashed, white, ultra thick] (-7,0.4)--(7,0.4);
\draw [loosely dashed, white, ultra thick] (-7,-0.9)--(7,-0.9);
\draw [white, ultra thick] (-7,-2.2)--(7,-2.2);

\pic[scale=0.25] at (-2,-1.55){car_top={blue}};

\draw[black,very thick, 
  postaction={
    decorate,
    decoration={
      markings,
      mark=between positions 0.15 and 1.0 step 0.275 with
      {
        \pic[transform shape,scale=0.25, opacity=0.45] {car_top={red}} ;
      }
    }
  }
]
(0,-1.55) -- (4,-1.55);

\pic[scale=0.25] at (0,-1.55){car_top={red}};

\end{tikzpicture}
    }
    \caption{Scenario Type A Concept—the ego vehicle is placed behind a decelerating challenger vehicle in the same lane.}
    \label{fig:scenario_a}
\end{figure}

\begin{figure} [h]
    \centering
    \resizebox{\linewidth}{!}{%
        \usetikzlibrary{decorations.markings}

\tikzset{
    car_top/.pic={
\filldraw [black, fill=#1, very thin] plot[smooth, black, tension=.7, very thin] 
	coordinates {(2.3248,0) (2.2967,0.4058) (2.0941,0.7925)(1.7847,0.9958) (1.1874,1.0261) (0.9497,1.0108)}
	.. controls (0.9124,1.143) and (0.8648,1.2352) .. (0.7463,1.1888) .. controls (0.7628,1.0703) and (0.7832,1.0089) .. (0.8202,0.952) .. controls (0.6029,0.9902) and (0.1582,1.0079) .. (-0.2536,0.9816) .. controls (-0.3325,1.0605) and (-0.4289,1.0459) .. (-0.5136,0.9787) .. controls (-0.6947,0.9904) and (-0.8232,0.9758) .. (-0.94,0.9524)
	plot[smooth, tension=.7,very thin] coordinates {(-0.94,0.9524)(-1.1951,1.0129) (-1.8081,0.9951) (-2.1686,0.8328) (-2.3894,0.4889) (-2.4286,0) };

\fill [fill=#1,very thin](-2.4286,0)--(-0.94,0.9524)--(2.3248,0)--(-0.94,-0.9524);

\filldraw  [black, fill=#1, very thin] plot[smooth, black, tension=.7,very thin] 
	coordinates {(2.3248,0) (2.2967,-0.4058) (2.0941,-0.7925)(1.7847,-0.9958) (1.1874,-1.0261) (0.9497,-1.0108)}
	.. controls (0.9124,-1.143) and (0.8648,-1.2352) .. (0.7463,-1.1888) .. controls (0.7628,-1.0703) and (0.7832,-1.0089) .. (0.8202,-0.952) .. controls (0.6029,-0.9902) and (0.1582,-1.0079) .. (-0.2536,-0.9816) .. controls (-0.3325,-1.0605) and (-0.4289,-1.0459) .. (-0.5136,-0.9787) .. controls (-0.6947,-0.9904) and (-0.8232,-0.9758) .. (-0.94,-0.9524)
	plot[smooth, tension=.7,very thin] coordinates {(-0.94,-0.9524)(-1.1951,-1.0129) (-1.8081,-0.9951) (-2.1686,-0.8328) (-2.3894,-0.4889) (-2.4286,0) };

\draw [smooth cycle, black, fill=darkgray,very thin](0.52,0.75) .. controls (0.7,0.78) and (1,0.82) .. (1.2,0.85) .. controls (1.7,0.5) and (1.7,-0.5) .. (1.2,-0.85) .. controls (1,-0.82) and (0.7,-0.78) .. (0.52,-0.75) .. controls (0.7,-0.25) and (0.7,0.25) .. (0.52,0.75);
\draw [smooth cycle, black, fill=darkgray,very thin](-1.06,0.67) .. controls (-1.33,0.68) and (-1.57,0.67) .. (-1.8,0.68) .. controls (-2.1,0.27) and (-2.1,-0.27) .. (-1.8,-0.68) .. controls (-1.57,-0.67) and (-1.33,-0.68) .. (-1.06,-0.67) .. controls (-1.16,-0.3) and (-1.16,0.3) .. (-1.06,0.67);
}}

\begin{tikzpicture}
\fill [gray] (-7,-2.3) rectangle (7,1.8);
\draw [white, ultra thick] (-7,1.7)--(7,1.7);
\draw [loosely dashed, white, ultra thick] (-7,0.4)--(7,0.4);
\draw [loosely dashed, white, ultra thick] (-7,-0.9)--(7,-0.9);
\draw [white, ultra thick] (-7,-2.2)--(7,-2.2);

\pic[scale=0.25] at (-4,-1.55){car_top={blue}};

\draw[black,very thick, 
  postaction={
    decorate,
    decoration={
      markings,
      mark=between positions 0.0 and 0.99 step 0.19 with
      {
        \pic[transform shape,scale=0.25, opacity=0.45] {car_top={red}} ;
      }
    }
  }
]
(-1.5,-1.55) 
.. controls (0.5,-1.55) and (1,-0.4) .. (2,-0.3)
.. controls (3,-0.2) and (4,-0.22) .. (4.6,-0.24)
.. controls (4.8,-0.245) and (5,-0.255) .. (5,-0.25);

\pic[scale=0.25] at (-1.5,-1.55){car_top={red}};

\pic[scale=0.25] at (1.5,-1.55){car_top={red}};

\draw[black,very thick, 
  postaction={
    decorate,
    decoration={
      markings,
      mark=between positions 0.15 and 1.0 step 0.275 with
      {
        \pic[transform shape,scale=0.25, opacity=0.45] {car_top={red}} ;
      }
    }
  }
]
(1.5,-1.55) -- (4.5,-1.55);

\pic[scale=0.25] at (1.5,-1.55){car_top={red}};


\end{tikzpicture}
    }
    \caption{Scenario Type Cutout Concept—two vehicles are placed ahead of the ego vehicle. The dynamic occlusion vehicle performs a cutout maneuver, exposing the ego vehicle to a rapidly decelerating vehicle in front. }
    \label{fig:scenario_c}
\end{figure}
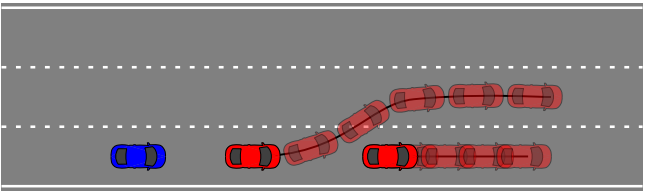

Synthetically generated scenarios are designed to present the ego vehicle with challenging situations where a collision is imminent unless the vehicle takes appropriate actions. Three types of synthetic scenarios were generated, inspired by the UN Regulations No. 157 for automated lane-keeping systems' critical scenarios \cite{UNECE2023}. The scenarios designed in this research are types A (Fig. \ref{fig:scenario_a}), B (Fig. \ref{fig:scenario_b}) and Cutout (Fig. \ref{fig:scenario_c}). The names Type A and B follow the convention from Weber et al. \cite{weber2019framework}.

\begin{figure} [h]
    \centering
    \resizebox{\linewidth}{!}{%
        \usetikzlibrary{decorations.markings}

\tikzset{
    car_top/.pic={
\filldraw [black, fill=#1, very thin] plot[smooth, black, tension=.7, very thin] 
	coordinates {(2.3248,0) (2.2967,0.4058) (2.0941,0.7925)(1.7847,0.9958) (1.1874,1.0261) (0.9497,1.0108)}
	.. controls (0.9124,1.143) and (0.8648,1.2352) .. (0.7463,1.1888) .. controls (0.7628,1.0703) and (0.7832,1.0089) .. (0.8202,0.952) .. controls (0.6029,0.9902) and (0.1582,1.0079) .. (-0.2536,0.9816) .. controls (-0.3325,1.0605) and (-0.4289,1.0459) .. (-0.5136,0.9787) .. controls (-0.6947,0.9904) and (-0.8232,0.9758) .. (-0.94,0.9524)
	plot[smooth, tension=.7,very thin] coordinates {(-0.94,0.9524)(-1.1951,1.0129) (-1.8081,0.9951) (-2.1686,0.8328) (-2.3894,0.4889) (-2.4286,0) };

\fill [fill=#1,very thin](-2.4286,0)--(-0.94,0.9524)--(2.3248,0)--(-0.94,-0.9524);

\filldraw  [black, fill=#1, very thin] plot[smooth, black, tension=.7,very thin] 
	coordinates {(2.3248,0) (2.2967,-0.4058) (2.0941,-0.7925)(1.7847,-0.9958) (1.1874,-1.0261) (0.9497,-1.0108)}
	.. controls (0.9124,-1.143) and (0.8648,-1.2352) .. (0.7463,-1.1888) .. controls (0.7628,-1.0703) and (0.7832,-1.0089) .. (0.8202,-0.952) .. controls (0.6029,-0.9902) and (0.1582,-1.0079) .. (-0.2536,-0.9816) .. controls (-0.3325,-1.0605) and (-0.4289,-1.0459) .. (-0.5136,-0.9787) .. controls (-0.6947,-0.9904) and (-0.8232,-0.9758) .. (-0.94,-0.9524)
	plot[smooth, tension=.7,very thin] coordinates {(-0.94,-0.9524)(-1.1951,-1.0129) (-1.8081,-0.9951) (-2.1686,-0.8328) (-2.3894,-0.4889) (-2.4286,0) };

\draw [smooth cycle, black, fill=darkgray,very thin](0.52,0.75) .. controls (0.7,0.78) and (1,0.82) .. (1.2,0.85) .. controls (1.7,0.5) and (1.7,-0.5) .. (1.2,-0.85) .. controls (1,-0.82) and (0.7,-0.78) .. (0.52,-0.75) .. controls (0.7,-0.25) and (0.7,0.25) .. (0.52,0.75);
\draw [smooth cycle, black, fill=darkgray,very thin](-1.06,0.67) .. controls (-1.33,0.68) and (-1.57,0.67) .. (-1.8,0.68) .. controls (-2.1,0.27) and (-2.1,-0.27) .. (-1.8,-0.68) .. controls (-1.57,-0.67) and (-1.33,-0.68) .. (-1.06,-0.67) .. controls (-1.16,-0.3) and (-1.16,0.3) .. (-1.06,0.67);
}}

\begin{tikzpicture}

\fill [gray] (-7,-2.3) rectangle (7,1.8);
\draw [white, ultra thick] (-7,1.7)--(7,1.7);
\draw [loosely dashed, white, ultra thick] (-7,0.4)--(7,0.4);
\draw [loosely dashed, white, ultra thick] (-7,-0.9)--(7,-0.9);
\draw [white, ultra thick] (-7,-2.2)--(7,-2.2);

\pic[scale=0.25] at (-2,-1.55){car_top={blue}};



\draw[
    black, very thick,
    postaction={
      decorate,
      decoration={
        markings,
        mark=between positions 0.15 and 0.95 step 0.19 with
          {\pic[transform shape, scale=0.25, opacity=0.45]{car_top={red}};}
      }
    }]
    (-3,-0.25)
  .. controls (-1.154,-0.25)  and (-0.692,-1.40) .. (0.231,-1.50)
  .. controls ( 1.154,-1.60)  and ( 2.077,-1.58) .. (2.631,-1.56)
  .. controls ( 2.815,-1.555) and ( 3.000,-1.545) .. (3.000,-1.55);
    
\pic[scale=0.25] at (-3.2,-0.25){car_top={red}};
    
\end{tikzpicture}
    }
    \caption{Scenario Type B Concept—the challenger vehicle is positioned in an adjacent lane and performs a cut-in maneuver in front of the ego vehicle, followed by deceleration.}
    \label{fig:scenario_b}
\end{figure}
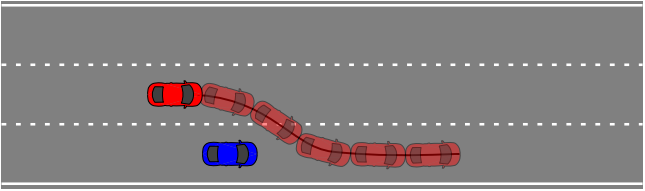

All scenarios were verified using the CommonRoad drivability checker \cite{pek2020commonroad}, ensuring they were free from collisions and physically feasible.

\subsubsection{Real Road Extracted Scenarios}
In addition to synthetic scenarios, real-road scenarios extracted from datasets like highD \cite{highDdataset} are used for validation. These scenarios were converted into XML format using the CommonRoad dataset converter \cite{lin2023automatic}, making them compatible with the simulator. Real-road scenarios provide a naturalistic component to the evaluation process and test the generalization ability of the model on unseen scenarios.

\subsection{Hierarchical Policy}
To address credit assignment issues associated with direct control in continuous space, we apply HRL. The high level agent selects discrete options, representing strategic driving intents, while the SAD-pilot handles low-level trajectory execution.
\noindent
The lateral action space includes:
\(a_\mathrm{{right}}\) (right lane change), \(a_\mathrm{{left}}\), and \(a_\mathrm{{center}}\) (keep lane).
\noindent
The longitudinal space includes:
\(a_\mathrm{{accelerate}}\), \(a_\mathrm{{brake}}\), \(a_\mathrm{{maintain}}\), and \(a_\mathrm{{hard\ brake}}\)

This abstraction improves learning stability and simplifies policy optimization.

\subsection{Shielding}
As a safety layer, the shield monitors the ego vehicle's state and stops the action if a high-risk action is detected (e.g., collision or going off the road). This ensures that the agent learns within safe and feasible bounds.

\subsection{Scenario Selection}
\label{sc:scenario_selection}

Not all scenarios are suitable for the HRL setup. Therefore, scenarios that require lane changes with insufficient decision time (under 6.5 seconds) are filtered out. This threshold considers the time required by the SAD-pilot to feasibly complete a lane change maneuver. The filtering ensures that all selected scenarios are suitable for hierarchical policy.

\subsection{Implementation}
\subsubsection{Simulation Setup}
The \gls{sad-rl} environment was implemented using FARAMA Gymnasium \cite{towers2024gymnasium} abstraction in addition to using the simulation and SAD-Pilot in the reset and step functions. The environment uses sparse reward system that rewards the agent if it reaches the goal and penalizes it otherwise.

The observation space in the simulation includes vehicle speed, distance to other vehicles, lane position, and surrounding 6 vehicles' distances and relative velocities.

\subsubsection{RL Agent}
The core learning algorithm is Advantage Actor-Critic (A2C), chosen for its balance of stability of learning and performance. A2C uses an actor-critic architecture which learns both a value function and a policy. It was implemented using the MultiInputPolicy architecture from Stable-Baselines3 \cite{raffin2021stable}, enabling it to process the structured observation space with multiple input components.

\section{Experimental Results}
\label{sc:experiments}
This section presents the evaluation setup and experimental results, focusing on the criteria: safety, learning efficiency, and generalizability. The results include both synthetic and real-road scenarios as well as baseline comparisons and ablation studies.

\subsection{Evaluation Framework}
The framework for evaluation adheres to the three metrics: \textit{safety}, \textit{learning efficiency}, and \textit{generalizability}.

\subsubsection{Safety}
Safety is quantified by analyzing the distribution of termination reasons at the end of each episode. The percentage of each termination reason $T_i$ is calculated as $P(T_i) = \frac{1}{M} \sum_{k=1}^{M} T_{i,k} \times 100$, where $M$ is the total number of evaluation episodes, and $T_{i,k}$ indicates whether reason $T_i$ caused termination in episode $k$. These metrics are visualized as moving averages to track safety trends during training.

\subsubsection{Efficiency of Learning}
Learning efficiency is assessed using moving average rates of termination reasons across episodes. The performance improvement is reflected in increasing Goal-Reaching rate and decreasing other termination reasons rates.

\subsubsection{Generalizability}
Generalizability is ensured by using distinct datasets for training and testing. This is tested with synthetic and real-road scenarios, using cross-validation between different scenario types.

\subsection{Initial Experiments}
We conducted training on both synthetic and real-road scenarios. Four candidate agents (DQN, QR-DQN, PPO, and A2C) were used for training. 

\subsubsection{Experiments Datasets}
Two sources of scenarios (synthetic and real-road) are used with distinct training-testing splits. The synthetic scenarios include 3000 training and 600 testing scenarios that includes 1000 training scenario from each type (A, B, and Cutout) as well as 200 testing scenarios from each type. For real-road scenarios, 300 training and 50 testing scenarios are used.

\subsubsection{Over-fitting Test} 
\label{sssection:overfitting}
This test was conducted to check the basic sanity of the 4 algorithms to show if there is a pattern of learning and to compare it later with the results in the ablation study. Agents were trained on a simple Type A scenario, where only one vehicle decelerates towards the ego vehicle. The training that lasted 200k steps for each algorithm resulted in:

\begin{table}[h]
\centering
\begin{tabular}{|l|c|c|}
\hline
\textbf{Algorithm} & \textbf{Goal-Reaching Rate} & \textbf{Collision Rate} \\
\hline
DQN & $< 45\%$ & $> 55\%$ \\
QR-DQN & $< 1\%$ & $> 90\%$ \\
PPO & $< 1\%$ & $> 95\%$ \\
A2C Fig. \ref{fig:termination_reasons_overfitting_a2c} & $> 95\%$ & $< 5\%$ \\
\hline
\end{tabular}
\caption{Performance comparison of different RL algorithms}
\label{tab:rl_comparison}
\end{table}

The results only show Goal-Reaching and Collision rates as the other termination reasons were negligible. The results show that A2C is the only algorithm that could show a concrete learning pattern on this one scenario. Therefore, it was selected for further experimentation.
\begin{figure}[h]
    \centering
    \includegraphics[width=0.48\textwidth]{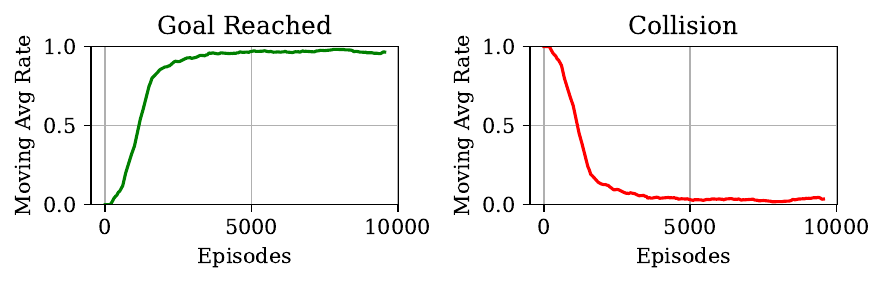}
    \caption{Training performance of the A2C agent on a single Type A scenario. The left plot shows a steady increase in goal-reaching rate, while the right plot indicates a significant reduction in collision rate over 10,000 episodes.}
    \label{fig:termination_reasons_overfitting_a2c}
\end{figure}

\subsubsection{Evaluation of A2C Policies Across Test Conditions}

To assess the generalization ability of various A2C policy variants, we define the goal-reaching rate $G(\pi_i, t) \in [0, 100]$ as the percentage of episodes in which agent $\pi_i$ successfully reaches the goal in test scenario $t \in \mathcal{T}$, where $\mathcal{T} = \{\text{HighD}, \text{A}, \text{B}, \text{Cutout}\}$.

Each policy $\pi_i$ is defined by its training configuration:

\begin{itemize}
\item $\pi_{\text{maintain}}$: Maintain lateral and longitudinal action (baseline).
\item $\pi_{\text{rand}}$: Random untrained stochastic policy.
\item $\pi_A$: A2C trained on 1000 scenarios of type A only.
\item $\pi_B$: A2C trained on 1000 scenarios of type B only.
\item $\pi_{\text{cut}}$: A2C trained on 1000 scenarios of type Cutout only.
\item $\pi_{\text{full}}$: A2C trained on 3000 scenarios (A, B, and Cutout).
\item $\pi_{\text{highD}}$: A2C trained on 300 real-road scenarios from the highD dataset.
\item $\pi_{\text{wes}}$: A2C trained on 3000 scenarios (A, B, and Cutout), including ``easy" scenarios that are solvable by the maintain policy.
\item $\pi_{\text{wes+highD}}$: A2C trained on the full With Easy Scenarios (WES) set (3000 scenarios) + 300 highD scenarios.
\end{itemize}

\vspace{0.5em}
\noindent The goal-reaching results $G(\pi_i, t)$ are summarized in Table~\ref{tab:test_goals}.

\begin{table}[h]
\centering
\caption{Goal-reaching rate $G(\pi_i, t)$ (\%) for each A2C agent $\pi_i$ across test scenarios $t \in \mathcal{T}$}
\label{tab:test_goals}
\resizebox{\linewidth}{!}{
\begin{tabular}{lcccc}
\toprule
\textbf{Policy $\pi_i$} & $G(\pi_i, \text{HighD})$ & $G(\pi_i, \text{A})$ & $G(\pi_i, \text{B})$ & $G(\pi_i, \text{Cutout})$ \\
\midrule
$\pi_{\text{maintain}}$ & 92.0 & 0.0 & 0.0 & 0.0 \\
$\pi_{\text{rand}}$ & 64.0 & 53.0 & 46.0 & 24.5 \\
$\pi_A$ & 70.0 & \textbf{95.5} & 57.5 & 56.5 \\
$\pi_B$ & 68.0 & 62.5 & \textbf{69.5} & 25.5 \\
$\pi_{\text{cut}}$ & 68.0 & 62.5 & 56.0 & \textbf{61.5} \\
$\pi_{\text{full}}$ & 70.0 & 74.5 & 65.5 & 54.0 \\
$\pi_{\text{highD}}$ & \textbf{88.0} & 52.0 & 61.5 & 16.0 \\
$\pi_{\text{wes}}$ & 76.0 & 86.0 & 68.0 & 60.0 \\
$\pi_{\text{wes+highD}}$ & \textbf{82.0} & \textbf{94.5} & \textbf{69.5} & \textbf{75.0} \\
\bottomrule
\end{tabular}
}
\end{table}

\noindent The results demonstrate that agents trained on single scenario types (e.g., $\pi_A$) show strong in-domain performance but lack generalization. The hybrid model $\pi_{\text{wes+highD}}$, trained on both easy synthetic and real-road data, achieves  high performance across all test cases, with $G(\pi_{\text{wes+highD}}, t) > 69\%$ for all $t \in \mathcal{T}$, making it the most robust policy overall.

\section{Discussion of Results}
\label{sc:discussion}
In this section, the results will be analyzed in addition to comparison studies to baseline algorithms and ablation studies to evaluate using HRL and scenario-based environment individually to asses their contribution.
\subsection{Full-Scale Training Across Agents}
\begin{figure}[h]
    \centering
    \includegraphics[width=0.48\textwidth]{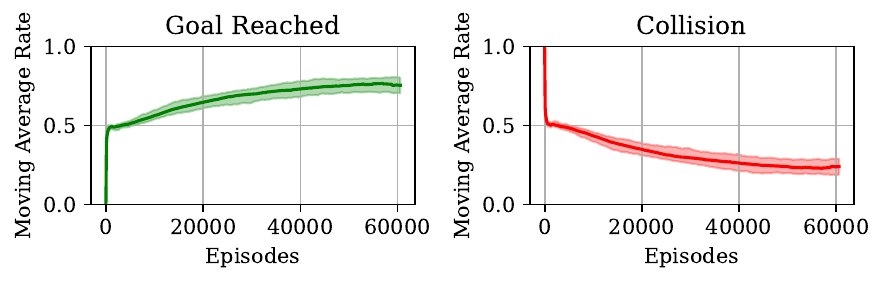}
    \caption{Moving average rate of training A2C agent for 1M timesteps with 10 different seeds.}
    \label{fig:a2c_curve}
\end{figure}

\begin{figure}[h]
    \centering
    \includegraphics[width=0.48\textwidth]{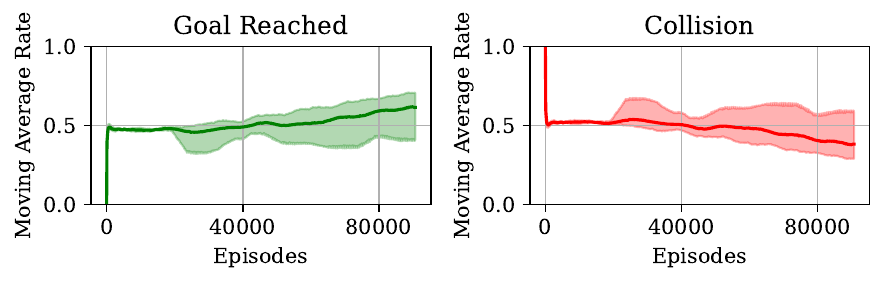}
    \caption{Moving average rate of training PPO agent for 1M timesteps with 10 different seeds.}
    \label{fig:ppo_curve}
\end{figure}

\begin{figure}[h]
    \centering
    \includegraphics[width=0.48\textwidth]{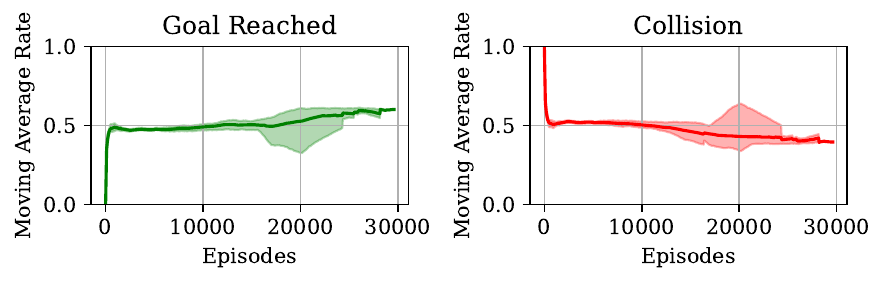}
    \caption{Moving average rate of training QR-DQN agent for 1M timesteps with 10 different seeds.}
    \label{fig:qr_dqn_curve}
\end{figure}

\begin{figure}[h]
    \centering
    \includegraphics[width=0.48\textwidth]{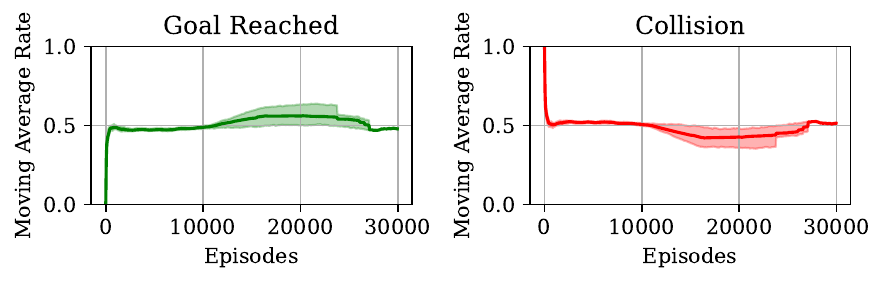}
    \caption{Moving average rate of training DQN agent for 1M timesteps with 10 different seeds.}
    \label{fig:dqn_curve}
\end{figure}

To evaluate generalizability and consistency under realistic training. We trained the four agents on the experiment dataset including 3300 training scenarios and 650 testing scenarios from both real-road and synthetic scenarios. Each algorithm was trained for 1 Million timesteps using 10 different seeds. Figures \ref{fig:a2c_curve} to \ref{fig:dqn_curve} present the training performance of each agent over the number of episodes. The number of episodes vary depending on how quickly the agent completes episodes (e.g. PPO is able to finish more episodes in 1 Million timesteps than the rest of the algorithms). The results show that DQN and A2C had the most consistent results while PPO and QR-DQN showed inconsistent results through different seeds. This shows that the \gls{sad-rl}  setup can accommodate different algorithms as well as showing that A2C was consistently the most successful agent in learning the scenarios.


\subsection{Ablation Studies}
\label{sc:ablation}
To assess the contributions of the two key design choices in this research HRL and challenging scenarios, we conduct two ablation studies. 

\subsubsection{Effect of HRL}
\begin{figure}[h]
    \centering
    \includegraphics[width=0.48\textwidth]{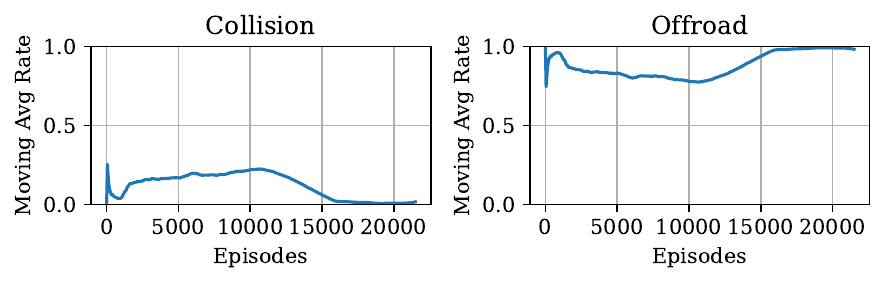}
    \caption{Training performance of A2C on one episode without HRL for 1M timesteps.}
    \label{fig:termination_reasons_overfitting_a2c_no_hrl}
\end{figure}

This ablation explores whether HRL and its associated lateral shield can significantly impact learning. In this setup, we revert the agent back to non-HRL version, operating in continuous action space (removing SAD-Pilot), and disabling lateral shielding due to its infeasibility in this setup. Only A2C is evaluated in this case as the best performing algorithm throughout the experiments as well as its compatibility with a continuous action space.
The agent was trained for overfitting on the same simple scenario trained in
section \ref{sssection:overfitting} but for 1 Million timesteps this time to give more chance to the agent. As seen in Fig. \ref{fig:termination_reasons_overfitting_a2c_no_hrl}, A2C was able to reduce its Collision rate, however the offroad termination rate approached 100\% by the end of the training. The agent failed to reach the goal even once. These results demonstrate that, without HRL, agents are unable to show robust driving behavior, which highlights the importance of HRL in this task.

The first ablation study focuses on the feasibility of using hierarchical policy with reinforcement learning. For this study, the use of HRL was completely ablated, returning the problem back to its original state that used a continuous action space. In addition, the use of the lateral shield was also ablated as it's not possible without the use of HRL. In this state, the agent has to pass two float numbers to the simulation: steering velocity and acceleration. The algorithms tested in this case are PPO and A2C. The reason for using these two out of the four algorithms used earlier is that value-based algorithms like DQN and QR-DQN are generally not compatible with continuous action space problems. 


\subsubsection{Effect of Challenging Scenarios}
To isolate the impact of challenging synthetic scenarios, we evaluate the generalization performance of the agent $\pi_{\text{highD}}$, which is trained solely on HighD, consisting of 300 real-road scenarios, for 1 million timesteps. The goal-reaching rates $G(\pi_{\text{highD}}, t)$ are computed for each $t \in \{\text{A}, \text{B}, \text{Cutout}\}$ and compared against other policies (see Table~\ref{tab:test_goals}).

While $\pi_{\text{highD}}$ achieves strong in-domain performance with $G(\pi_{\text{highD}}, \text{HighD}) = 88.0\%$, its performance degrades significantly under out-of-distribution test scenarios. Interestingly, it slightly outperforms some single-type agents on $G(\pi_{\text{highD}}, \text{B}) = 61.5\%$. These results indicate that exposure to real-world scenarios alone is insufficient for robust generalization across diverse driving behaviors. Instead, a combination of structured adversarial training data and naturalistic real-road experiences is necessary to achieve strong cross-domain performance.

\subsection{Summary of Key Findings}
\begin{enumerate}
    \item HRL is critical for learning in \gls{sad-rl}  setup, it allows the agent to discover goal-directed policies and ensures lateral safety
    \item Scenario diversity is essential for generalization; models trained solely on real-road scenarios lacked robustness when tested with challenging or unseen setups.
    \item Training across different random seeds revealed that A2C and DQN are more consistent learners, especially when supported by HRL and diverse training data
\end{enumerate}

\section{Conclusion}
In this work, we presented the \gls{sad-rl} framework for training automated driving decision-making algorithms using \gls{rl}.
The framework leverages HRL to improve learning efficicency and safety and implements a scenario-based approach for the training process.
This process provides more control on the experience used for learning than random traffic.
The results of our experiments indicate that the \gls{sad-rl} framework is a promising approach to efficently develop safe decision-making algorithms for complex driving task.
Our ablation studies show that both, HRL and the scenario-based approach, are essential features of our framework.

Future work will focus on expanding the range of scenarios and tasks in the \gls{sad-rl} environment, incorporating more complex urban driving situations, increasing scenario diversity, and further refining the HRL architecture to handle additional challenges such as pedestrian interaction and complex traffic rules.

\bibliographystyle{IEEEtran}
\bibliography{bibtex/bib/IEEEexample}

\begin{thebibliography}{10}
\providecommand{\url}[1]{#1}
\csname url@samestyle\endcsname
\providecommand{\newblock}{\relax}
\providecommand{\bibinfo}[2]{#2}
\providecommand{\BIBentrySTDinterwordspacing}{\spaceskip=0pt\relax}
\providecommand{\BIBentryALTinterwordstretchfactor}{4}
\providecommand{\BIBentryALTinterwordspacing}{\spaceskip=\fontdimen2\font plus
\BIBentryALTinterwordstretchfactor\fontdimen3\font minus \fontdimen4\font\relax}
\providecommand{\BIBforeignlanguage}[2]{{%
\expandafter\ifx\csname l@#1\endcsname\relax
\typeout{** WARNING: IEEEtran.bst: No hyphenation pattern has been}%
\typeout{** loaded for the language `#1'. Using the pattern for}%
\typeout{** the default language instead.}%
\else
\language=\csname l@#1\endcsname
\fi
#2}}
\providecommand{\BIBdecl}{\relax}
\BIBdecl

\bibitem{10.3389/fhumd.2021.669030}
\BIBentryALTinterwordspacing
F.~Hartwich, C.~Hollander, D.~Johannmeyer, and J.~F. Krems, ``Improving passenger experience and trust in automated vehicles through user-adaptive hmis: “the more the better” does not apply to everyone,'' \emph{Frontiers in Human Dynamics}, vol.~3, 2021. [Online]. Available: \url{https://www.frontiersin.org/articles/10.3389/fhumd.2021.669030}
\BIBentrySTDinterwordspacing

\bibitem{kremer2020scenario}
\BIBentryALTinterwordspacing
M.~Kremer, S.~Christiaens, C.~Granrath \emph{et~al.}, ``Scenario- and model-based systems engineering for highly automated driving,'' \emph{ATZ Worldw}, vol. 122, pp. 16--21, 2020. [Online]. Available: \url{https://doi.org/10.1007/s38311-020-0330-x}
\BIBentrySTDinterwordspacing

\bibitem{silver2016mastering}
D.~Silver, A.~Huang, C.~J. Maddison, A.~Guez, L.~Sifre, G.~Van Den~Driessche, J.~Schrittwieser, I.~Antonoglou, V.~Panneershelvam, M.~Lanctot \emph{et~al.}, ``Mastering the game of go with deep neural networks and tree search,'' \emph{nature}, vol. 529, no. 7587, pp. 484--489, 2016.

\bibitem{9210154}
S.~Aradi, ``Survey of deep reinforcement learning for motion planning of autonomous vehicles,'' \emph{IEEE Transactions on Intelligent Transportation Systems}, vol.~23, no.~2, pp. 740--759, 2022.

\bibitem{liu2022combining}
T.~Liu, Q.~Liu, H.~Liu, and X.~Ren, ``Combining deep reinforcement learning with rule-based constraints for safe highway driving,'' in \emph{2022 China Automation Congress (CAC)}.\hskip 1em plus 0.5em minus 0.4em\relax IEEE, 2022, pp. 2785--2790.

\bibitem{wang2023efficient}
L.~Wang, J.~Liu, H.~Shao, W.~Wang, R.~Chen, Y.~Liu, and S.~L. Waslander, ``Efficient reinforcement learning for autonomous driving with parameterized skills and priors,'' \emph{arXiv preprint arXiv:2305.04412}, 2023.

\bibitem{lu2023imitation}
Y.~Lu, J.~Fu, G.~Tucker, X.~Pan, E.~Bronstein, R.~Roelofs, B.~Sapp, B.~White, A.~Faust, S.~Whiteson \emph{et~al.}, ``Imitation is not enough: Robustifying imitation with reinforcement learning for challenging driving scenarios,'' in \emph{2023 IEEE/RSJ International Conference on Intelligent Robots and Systems (IROS)}.\hskip 1em plus 0.5em minus 0.4em\relax IEEE, 2023, pp. 7553--7560.

\bibitem{fraichard1993dynamic}
T.~Fraichard, ``Dynamic trajectory planning with dynamic constraints: A'state-time space'approach,'' in \emph{Proceedings of 1993 IEEE/RSJ International Conference on Intelligent Robots and Systems (IROS'93)}, vol.~2.\hskip 1em plus 0.5em minus 0.4em\relax IEEE, 1993, pp. 1393--1400.

\bibitem{ajanovic2018search}
Z.~Ajanovic, B.~Lacevic, B.~Shyrokau, M.~Stolz, and M.~Horn, ``Search-based optimal motion planning for automated driving,'' in \emph{2018 IEEE/RSJ International Conference on Intelligent Robots and Systems (IROS)}.\hskip 1em plus 0.5em minus 0.4em\relax IEEE, 2018, pp. 4523--4530.

\bibitem{saraoglu2023minimax}
M.~Saraoglu, H.~Jiang, M.~Schirmer, {\.I}.~Mutlu, and K.~Janschek, ``A minimax-based decision-making approach for safe maneuver planning in automated driving,'' in \emph{2023 American Control Conference (ACC)}.\hskip 1em plus 0.5em minus 0.4em\relax IEEE, 2023, pp. 4683--4690.

\bibitem{pomerleau1988alvinn}
D.~A. Pomerleau, ``Alvinn: An autonomous land vehicle in a neural network,'' \emph{Advances in neural information processing systems}, vol.~1, 1988.

\bibitem{igl2022symphony}
M.~Igl, D.~Kim, A.~Kuefler, P.~Mougin, P.~Shah, K.~Shiarlis, D.~Anguelov, M.~Palatucci, B.~White, and S.~Whiteson, ``Symphony: Learning realistic and diverse agents for autonomous driving simulation,'' in \emph{2022 International Conference on Robotics and Automation (ICRA)}.\hskip 1em plus 0.5em minus 0.4em\relax IEEE, 2022, pp. 2445--2451.

\bibitem{alshiekh2018safe}
M.~Alshiekh, R.~Bloem, R.~Ehlers, B.~K{\"o}nighofer, S.~Niekum, and U.~Topcu, ``Safe reinforcement learning via shielding,'' in \emph{Proceedings of the AAAI conference on artificial intelligence}, vol.~32, no.~1, 2018.

\bibitem{vitelli2022safetynet}
M.~Vitelli, Y.~Chang, Y.~Ye, A.~Ferreira, M.~Wo{\l}czyk, B.~Osi{\'n}ski, M.~Niendorf, H.~Grimmett, Q.~Huang, A.~Jain \emph{et~al.}, ``Safetynet: Safe planning for real-world self-driving vehicles using machine-learned policies,'' in \emph{2022 International Conference on Robotics and Automation (ICRA)}.\hskip 1em plus 0.5em minus 0.4em\relax IEEE, 2022, pp. 897--904.

\bibitem{Althoff2017a}
M.~Althoff, M.~Koschi, and S.~Manzinger, ``Commonroad: Composable benchmarks for motion planning on roads,'' in \emph{Proc. of the IEEE Intelligent Vehicles Symposium}, 2017, pp. 719--726.

\bibitem{UNECE2023}
{United Nations Economic Commission for Europe}, ``Un regulation no. 157 - automated lane keeping systems (alks),'' \url{https://unece.org/sites/default/files/2023-12/R157e.pdf}, 2023, accessed: 2024-07-09.

\bibitem{weber2019framework}
H.~Weber, J.~Bock, J.~Klimke, C.~Roesener, J.~Hiller, R.~Krajewski, A.~Zlocki, and L.~Eckstein, ``A framework for definition of logical scenarios for safety assurance of automated driving,'' \emph{Traffic injury prevention}, vol.~20, no. sup1, pp. S65--S70, 2019.

\bibitem{pek2020commonroad}
C.~Pek, V.~Rusinov, S.~Manzinger, M.~C. {\"U}ste, and M.~Althoff, ``Commonroad drivability checker: Simplifying the development and validation of motion planning algorithms,'' in \emph{2020 IEEE intelligent vehicles symposium (IV)}.\hskip 1em plus 0.5em minus 0.4em\relax IEEE, 2020, pp. 1013--1020.

\bibitem{highDdataset}
R.~Krajewski, J.~Bock, L.~Kloeker, and L.~Eckstein, ``The highd dataset: A drone dataset of naturalistic vehicle trajectories on german highways for validation of highly automated driving systems,'' in \emph{2018 21st International Conference on Intelligent Transportation Systems (ITSC)}, 2018, pp. 2118--2125.

\bibitem{lin2023automatic}
Y.~Lin, M.~Ratzel, and M.~Althoff, ``Automatic traffic scenario conversion from openscenario to commonroad,'' \emph{arXiv preprint arXiv:2305.10080}, 2023.

\bibitem{towers2024gymnasium}
M.~Towers, A.~Kwiatkowski, J.~Terry, J.~U. Balis, G.~De~Cola, T.~Deleu, M.~Goul{\~a}o, A.~Kallinteris, M.~Krimmel, A.~KG \emph{et~al.}, ``Gymnasium: A standard interface for reinforcement learning environments,'' \emph{arXiv preprint arXiv:2407.17032}, 2024.

\bibitem{raffin2021stable}
A.~Raffin, A.~Hill, A.~Gleave, A.~Kanervisto, M.~Ernestus, and N.~Dormann, ``Stable-baselines3: Reliable reinforcement learning implementations,'' \emph{Journal of machine learning research}, vol.~22, no. 268, pp. 1--8, 2021.

\end{thebibliography}

\end{document}